\documentclass[10pt,twocolumn,letterpaper]{article}

\usepackage{cvpr}
\usepackage{times}
\usepackage{epsfig}
\usepackage{graphicx}
\usepackage{amsmath}
\usepackage{amssymb}

\usepackage{color,graphicx,amssymb,amsmath}
\usepackage{array}
\usepackage{balance}
\usepackage{subcaption}
\newcommand{\zheng}[1]{}
\def\mcL{\mathcal{L}}
\def\mcH{\mathcal{H}}

\usepackage[pagebackref=true,breaklinks=true,letterpaper=true,colorlinks,bookmarks=false]{hyperref}

 \cvprfinalcopy 

\def\cvprPaperID{2092} 

\ifcvprfinal\fi
\begin{document}

\title{Training Shallow and Thin Networks for Acceleration via Knowledge Distillation with Conditional Adversarial Networks}


\author{Zheng Xu\thanks{zuxh@cs.umd.edu}\\
University of Maryland\\ College Park \\
\and Yen-Chang Hsu \\
Georgia Institute of Technology\\ Atlanta\\
\and Jiawei Huang\\
Honda Research Institute\\ Mountain View\\
}

\maketitle

\begin{abstract}
There is an increasing interest on accelerating neural networks for real-time applications. We study the student-teacher strategy, in which a small and fast student network is trained with the auxiliary information learned from a large and accurate teacher network. We propose to use conditional adversarial networks to learn the loss function to transfer knowledge from teacher to student. The proposed method is particularly effective for relatively small student networks.  Moreover, experimental results  show the effect of network size when the modern networks are used as student. We empirically study the trade-off between inference time and classification accuracy, and provide suggestions on choosing a proper student network.
\end{abstract}

\section{Introduction}
Deep neural networks (DNNs) achieve massive success in artificial intelligence by substantially improving the state-of-the-art performance in various applications. For one of the core applications in computer vision, large-scale image classification \cite{russakovsky2015imagenet}, the accuracy reached by DNNs has become comparable to humans on several benchmark datasets. 
The recent progress towards such impressive accomplishment is largely driven by exploring deeper and wider network architectures. 
Despite the significant performance boost of modern DNNs \cite{he2016deep,zagoruyko2016wide,xie2016aggregated}, the heavy computation and memory cost of these deep and wide networks makes it difficult to directly deploy the trained networks on embedded systems for real-time applications. In the meantime, the demand for low cost networks is increasing for applications on mobile devices and autonomous cars.

Do DNNs really need to be deep and wide? Early theoretical studies suggest that shallow networks are powerful and can approximate arbitrary functions \cite{cybenko1989approximation,hornik1989multilayer}. More recent theoretical results show depth is indeed beneficial for the expressive capacity of networks \cite{eldan2016power,telgarsky2016benefits,liang2016deep,safran2017depth}. Moreover, the overparameterized and redundant networks, which can easily memorize and overfit the training data, surprisingly generalize well in practice \cite{zhang2016understanding,arpit2017closer}. 
Various explanations have been investigated, but the secret of deep and wide networks remains an open problem.

On the other hand, empirical studies suggest that the performance of shallow networks  can be improved by learning from large networks following the student-teacher strategy
\cite{buciluǎ2006model,ba2014deep,urban2016deep,hinton2015distilling}. 
In these approaches, the student networks are forced to mimic the output probability distribution of the teacher networks to transfer the knowledge embedded in the soft targets. The intuition is that the \emph{dark knowledge} \cite{hinton2015distilling}, which contains the relative probabilities of ``incorrect'' answers provided by deep and wide networks, is informative and representative.  
For example, we want to classify an image over the label set (dog, cat, car). Given an image of a dog, a good teacher network may mistakenly recognize it as cat with small probability, but should seldom recognize it as car; the soft target of output distribution over categories for this image, $(0.7,0.3,0)$, contains more information such as categorical correlation than the hard target of one-hot vector, $(1,0,0)$. Training is accomplished by minimizing a predetermined loss which measures similarity between student and teacher output, such as Kullback-Leibler (KL) divergence.

In previous studies,  shallow and wide student networks are trained by knowledge transfer, which potentially have more parameters than the deep teacher networks \cite{ba2014deep,urban2016deep};  ensemble of networks are used as teacher, and a student network with similar architecture and capacity can be trained \cite{hinton2015distilling}; particularly,  a small deep and thin network is trained to replace a shallow and wide network for acceleration \cite{romero2014fitnets}, given the best teacher at that time is the shallow and wide VGGNet \cite{simonyan2014very}. 
Since then, the design of network architecture has advanced. ResNet \cite{he2016deep} has significantly deepened the networks by introducing residual connections, and wide residual networks (WRNs) \cite{zagoruyko2016wide} suggest widening the networks leads to better performance. 
It is unclear whether the dark knowledge from the state-of-the-art networks based on residual connections, 
which are both deep and wide, can help train a shallow and/or thin network (also with residual connections) for acceleration. 

In this paper, we focus on improving the performance of a shallow and thin modern network (student) by learning from the dark knowledge of a deep and wide network (teacher). 
Both the student and teacher networks are convolutional neural networks (CNNs) with residual connections, and the student network is shallow and thin so that it can run much faster than the teacher network during inference. 
Instead of adopting the classic student-teacher strategy of forcing the output of a student network to exactly mimic the soft targets produced by a teacher network, we introduce conditional adversarial networks to transfer the dark knowledge from teacher to student. 
We empirically show that the loss learned by the adversarial training has the advantage over the predetermined loss in the student-teacher strategy, especially when the student network has relatively small capacity. 
%
%

\begin{figure}[t]
\centerline{
\includegraphics[width=\linewidth]{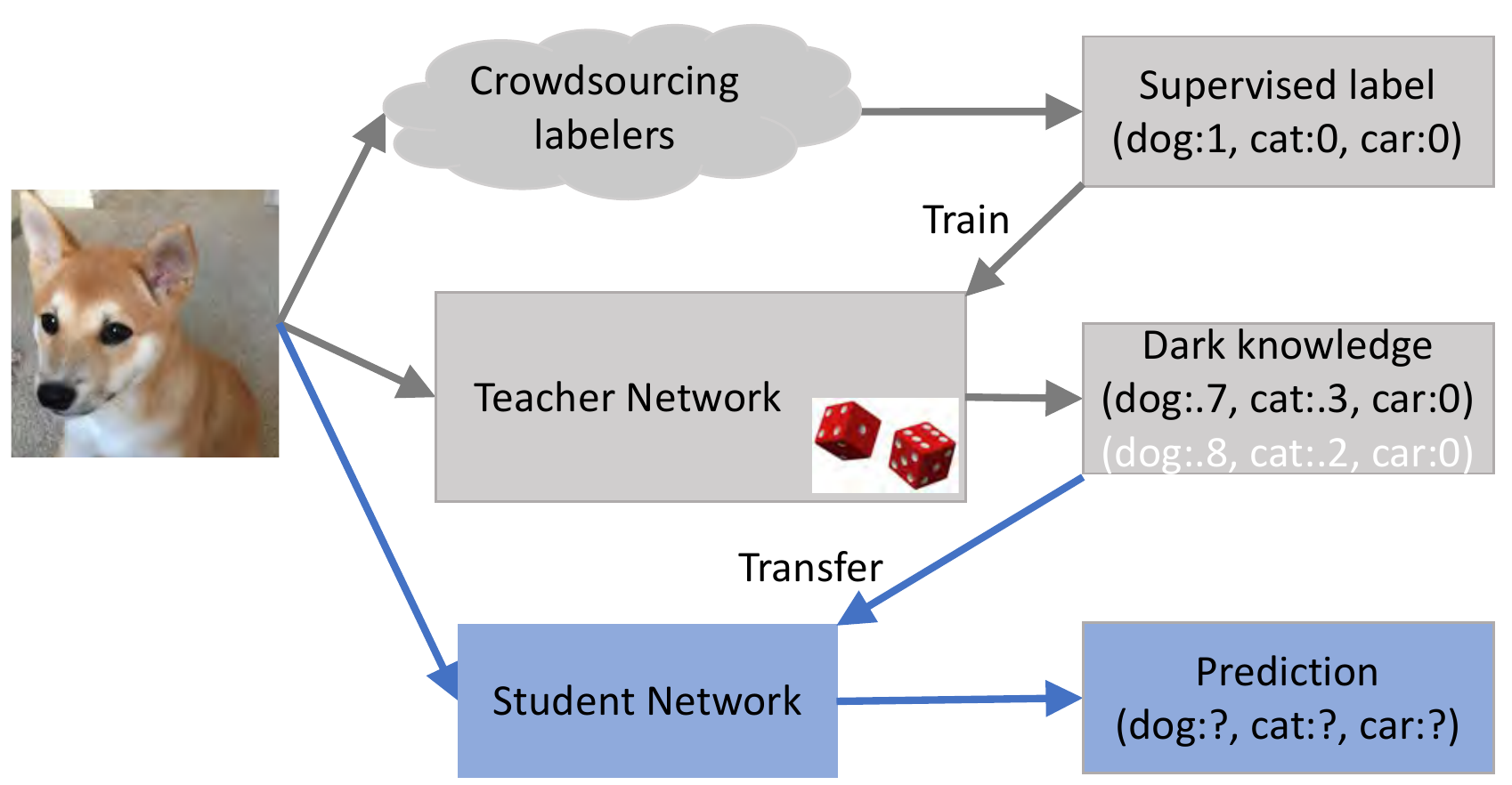}
}
\caption{
The motivation for our GAN-based student-teacher strategy: the soft targets produced by teacher network is more informative and the learned loss can transfer the multi-modal knowledge.}
\label{fig:motive}
\vspace{-0.5cm}
\end{figure}

Our learning loss approach is inspired by the recent success of conditional adversarial networks for  various image-to-image translation applications \cite{isola2016image}. 
We show that the generative adversarial nets (GANs) can benefit a task that is very different from image generation. 
In the student-teacher strategy, GAN can help preserve the multi-modal~\footnote{We explain the multi-modality with the previous example: the output distribution for a dog image can also be (0.8, 0.2, 0). In fact, there are infinite number of soft targets that can correctly predict the label.\zheng{still need to explain mutli-modal.}} nature  of the output distribution. 
It is not only unnecessary, but also difficult to force a student network to exactly mimic one of the soft targets (or the average/ensemble of several teacher networks), because the student has smaller capacity than the teacher. 
By introducing the discriminator as in GAN, the network automatically learns a good loss to transfer the correlation between classes, i.e., the dark knowledge from teacher, and also preserves the multi-modality. We summarize the motivation for our approach in Figure \ref{fig:motive}.

\section{Related work}
\textbf{Network acceleration} has gained increasing interest due to the growing needs of real-time applications in artificial intelligence. 
The techniques can be roughly divided into three categories: low precision, parameter pruning and factorization, and knowledge distillation. 
Low precision methods use limited number of bits to store and operate the network weights, and the extreme case is binary networks that only use 1-bit to represent each number \cite{rastegari2016xnor,li2017training}. 
The acceleration of these methods is somewhat conceptual because mainstream GPUs only have limited support for low precision computation. 
Networks can also be directly modified by pruning and factorizing the redundant weights, either as a post-processing step after training, or as a fine-tuning stage \cite{li2016pruning,howard2017mobilenets}. 
These methods often assume network weights are sparse or low rank, and aim to construct networks of similar architecture with reduced number of weights. Moreover, network pruning papers mostly report speedup indirectly measured in the number of basic operations, rather than by inference time directly.

\textbf{Knowledge distillation} is a principled approach to train small neural networks for acceleration. 
We slightly generalize the term \emph{knowledge distillation} to represent all methods that train student networks by transferring knowledge from teacher networks. 
Bucilua \etal \cite{buciluǎ2006model} pioneered this approach for model compression. Ba and Caruana\cite{ba2014deep}, and Urban \etal \cite{urban2016deep} trained shallow but wide student by learning from a deep teacher, which were not primarily designed for acceleration. 
Hinton \etal \cite{hinton2015distilling} generalized the previous methods by introducing a new metric between the output distribution of teacher and student, as well as a tuning parameter. 
Variants of knowledge distillation has also been applied to many different tasks, such as semantic segmentation \cite{ros2016training}, pedestrian detection \cite{shen2016teacher}, face recognition \cite{luo2016face}, metric learning\cite{chen2017darkrank},  reinforcement learning \cite{teh2017distral} and for regularization\cite{sau2016deep}.
A recent preprint \cite{kim2017transferring} presented promising preliminary results on CIFAR-10 by learning a small ResNet from a large ResNet. 
Another line of research focuses on transferring intermediate features instead of soft targets from teacher to student \cite{romero2014fitnets,wang2016deeply,zagoruyko2016paying,yimgift,huang2017like,zhou2017rocket,you2017learning}. 
Our approach is complementary to those methods by directly following \cite{hinton2015distilling} to design a new metric  between the output distribution of teacher and student, and adversarial networks are used to learn the metric to replace hand-engineering.

\textbf{Generative adversarial networks (GAN)} has been extensively studied over recent years since \cite{goodfellow2014generative}. GAN trains two neural networks, the generator and the discriminator, in an adversarial learning process that alternatively updates the two networks. We apply GAN in the conditional setting \cite{mirza2014conditional,isola2016image,reed2016generative,odena2016conditional}, where the generator is conditioned on input images.  Unlike previous works that focused on image generation, we aim at learning a loss function for knowledge distillation, which requires quite different architectural choices for our generator and discriminator.

\section{Learning loss for knowledge distillation}
In this section, we introduce the learning loss approach based on conditional adversarial networks. 
We start from a recap of modern network architectures (section \ref{sec:res}), and then describe the dark knowledge that can be transferred from teacher to student networks (section \ref{sec:dk}). 
The GAN-based approach for learning loss is detailed in section \ref{sec:gan}.

\subsection{Neural networks with residual connection}
\label{sec:res}
\begin{figure}[tbhp]
\centerline{
\includegraphics[width=0.9\linewidth]{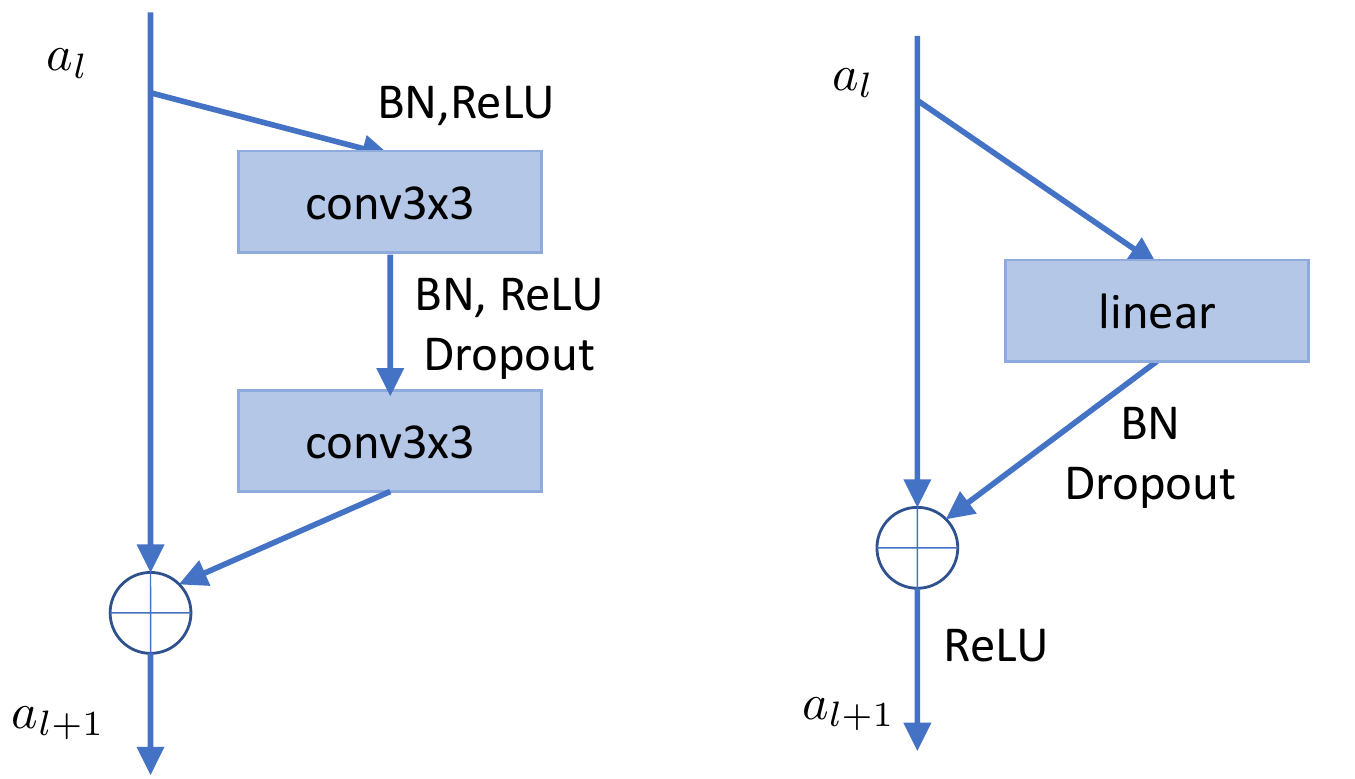}
}
\caption{\small Residual blocks for convolutional neural networks \cite{zagoruyko2016wide} (left) and multi-layer perceptron (right). $a_l$ represents the  output of the $l$th block. Each block is composed of batch normalization (BN), activation ReLU, weight layer, and dropout.}
\label{fig:res}
\end{figure}
The modern neural networks are built by stacking basic components. 
For computer vision tasks, residual blocks \cite{he2016deep,zagoruyko2016wide} are the basic components to build deep neural networks to achieve state-of-the-art performance. 
Both student and teacher networks in this paper are based on the residual convolutional blocks shown in Figure \ref{fig:res} (left). 
The first layer contains $16$ filters of $3 \times 3$ convolution, followed by a stack of $6n$ layers, which is 3 groups of $n$ residual blocks, and each block contains two convolution layers equipped with batch normalization \cite{ioffe2015batch}, ReLU \cite{krizhevsky2012imagenet} and dropout \cite{srivastava2014dropout}. 
The output feature map is subsampled twice, and the number of filters are doubled when subsampling, as shown in Table \ref{tab:res}.
After the last residual block is the global average pooling, and then fully-connected layer and softmax.
 In the following sections, the architecture of wide residual networks (WRNs) is denoted as WRN-$d$-$m$ following \cite{zagoruyko2016wide}\zheng{it follows from the paper, I think it also counts for average pooling and the other output stuff, not pretty sure though, don't want to make mistake on those minor stuff}, where the total depth is $d=6n+4$, and $m$ is the widen factor used to increase the number of filters in  each residual block.
Our teacher network is deep and wide WRN with large $d$ and $m$, while student network is shallow and thin WRN with small $d$ and $m$.

\begin{table}[tbhp]
\centering
\begin{tabular}{|c|c|c|c|}
\hline
& output size & \# layers & \# filters \\
\hline
\hline
group1 & 32 $\times$ 32 & 2n & 16m \\
\hline
group2 & 16 $\times$ 16 & 2n & 32m \\
\hline
group3 & 8 $\times$ 8 & 2n & 64m \\
\hline
\end{tabular}%
\caption{\small The stacked architecture of wide residual networks \cite{zagoruyko2016wide}. $n$ represents the number of residual blocks, $m$ represents the widen factor.}
\label{tab:res}%
\end{table}%

\subsection{Knowledge distillation}
\label{sec:dk}
The output of neural networks for image classification is a probability distribution over categories. 
The probability is generated by applying a softmax function over the output of the last fully connected layer, also known as \emph{logits}. 
The dimension of logits from student and teacher networks are both equal to the number of categories. 
Rich information is embedded in the output of a teacher network, and we can use logits to transfer the knowledge to student  network \cite{buciluǎ2006model,ba2014deep,urban2016deep,hinton2015distilling}. We review the method in \cite{hinton2015distilling}, which provides a metric between student and teacher logits that generalized previous methods for \emph{knowledge distillation.} We denote this work as KD for simplicity.

The logits vector generated by pre-trained teacher network for an input image $x_i, i=1,\ldots,N$ is represented by $t_i$, where the dimension of vector $t_i = (t_i^1, \ldots, t_i^C)$ is the number of categories $C$. 
We now consider training a student network $F$ to generate student logits $F(x_i)$. 
 By introducing a parameter called temperature $T$, the generalized softmax layer converts logits vector $t_i$ to probability distribution $q_i$,
\begin{equation}
M_T(t_i) = q_i, \text{ where } q_i^j = \frac{\exp(t_i^j/T)}{\sum_k \exp(t_i^k/T)}.
\end{equation}
where higher temperature $T$ produces softer probability over categories. 
The regular softmax for classification is a special case of the generalized softmax with $T=1$.

Hinton \etal  \cite{hinton2015distilling}\zheng{it looks weird to say 'KD' proposed sth. Authors look better} proposed to minimize the KL divergence between teacher and student output,
\begin{equation}
\mcL_{KD}(F, T) = \frac{1}{N} \sum_{i=1}^N \text{KL}(M_T(t_i) \|  M_T(F(x_i))). 
\end{equation}
It can be shown that when $T$ is very large, $\mcL_{KD}$ becomes the Euclidean distance between teacher and student logits, $\|t_i - F(x_i)\|^2_2$.

When the image-label pairs $\{x_i, l_i\}$ are provided, the cross-entropy loss for supervised training of a neural network can be represented as 
\begin{equation}
\mcL_{S}(F) = \frac{1}{N} \sum_{i=1}^N \mcH (l_i,  M_1(F(x_i))). \label{eq:ce}
\end{equation}
$\mcL_{S}$  is a commonly used loss for pure supervised learning in image classification from annotated data. 

Finally, Hinton \etal  \cite{hinton2015distilling} proposed to minimize the weighted sum of loss $\mcL_{KD}$ and loss $\mcL_{S}$ to train a student network, 
\begin{equation}
\mcL_1(F, T) = \frac{1}{2}  \mcL_{S} (F) + T^2 \mcL_{KD}(F, T). \label{eq:kdfinal}
\end{equation}

\subsection{Learning loss with adversarial networks}
\label{sec:gan}

\subsubsection{Overview}
\begin{figure}[tbhp]
\centerline{
\includegraphics[width=0.8\linewidth]{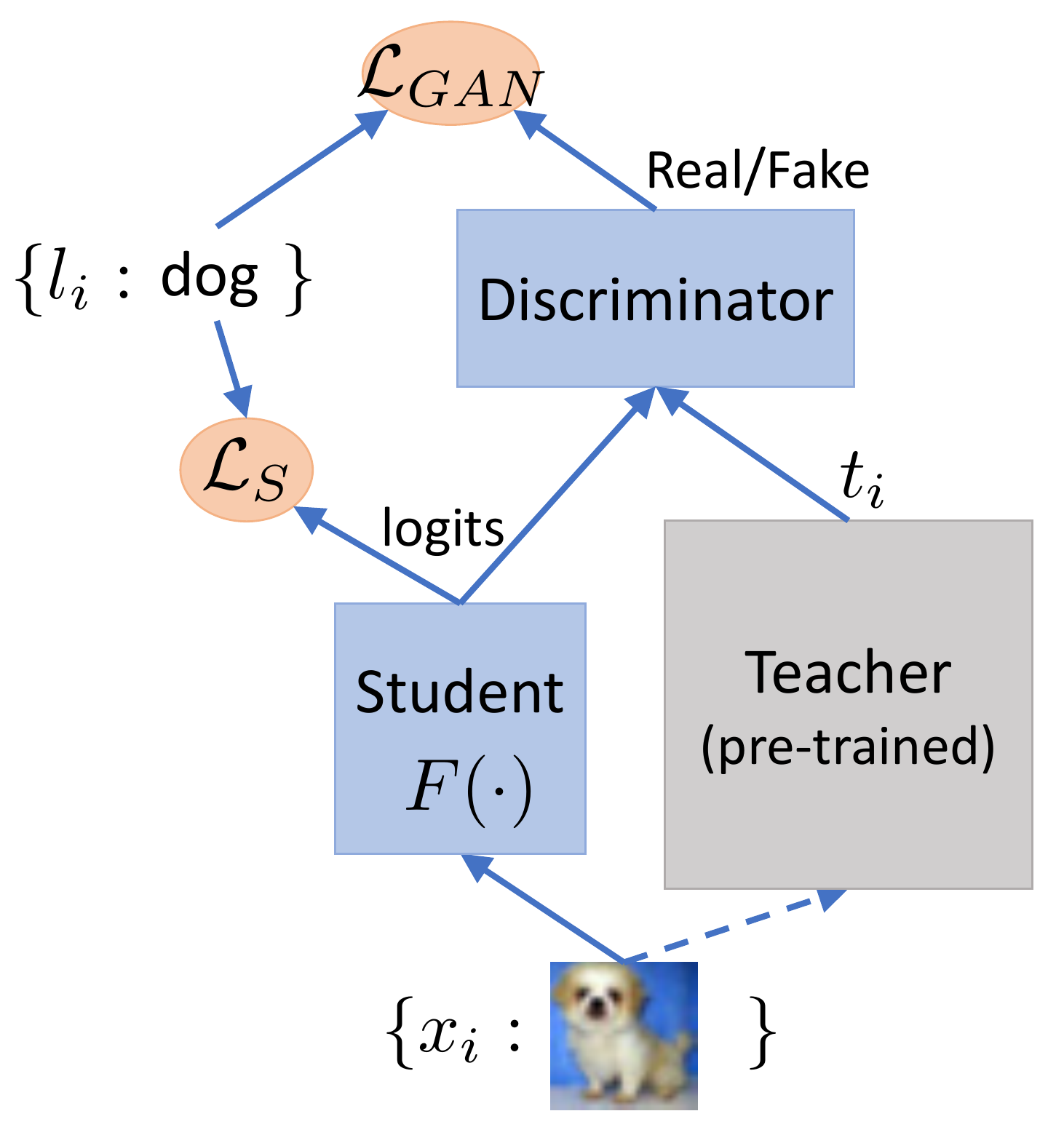}
}
\caption{
The GAN-based architecture to learn loss for knowledge distillation. The deep and wide teacher is pre-trained offline. The student network and discriminator are updated alternatively, where the discriminator aims to distinguish logits from student and teacher networks, and the student aims to fool the discriminator. Additional supervised loss is added for both student and discriminator.}
\label{fig:gan}
\end{figure}

The main idea of learning the  loss for transferring knowledge from teacher to student is depicted in Figure \ref{fig:gan}. 
Instead of forcing the student to exactly mimic the teacher by minimizing KL-divergence in $\mcL_1(F, T)$ of Equation \eqref{eq:kdfinal}, the knowledge is transferred from teacher to student through a discriminator in our GAN-based approach. 
This discriminator is trained to distinguish whether the output logits is from teacher or student network, while the student (the generator) is adversarially trained to fool the discriminator, i.e., output logits similar to the teacher logits so that the discriminator can not distinguish. 

There are several benefits of the proposed method. First,  the learned loss can be effective, as has already been demonstrated for several image to image translation tasks \cite{isola2016image}. Moreover, the GAN-based approach relieves the pain for hand-engineering the loss. 
Though the parameter tuning and hand-engineering of the loss is replaced by hand-engineering the discriminator networks in some sense, our empirical study shows that the performance is less sensitive to the discriminator architecture than the temperature parameter in knowledge distillation. 
The second benefit is closely related to the multi-modality of network output. Let us revisit the example of classifying a dog image from the label set (dog, cat, car). Both (0.7, 0.3, 0) and (0.8, 0.2, 0) are outputs can give correct prediction (dog), therefore it is not necessary to exactly mimic the output of one teacher network to achieve good student performance. Given the small capacity of the student network, it may not be able to exactly reproduce one particular output modality. The usage of discriminator relaxes the rigid coupling between student and teacher. The relative similarities between the categories can be captured by the discriminator trained from the multi-modal logits of teacher. Knowledge transferred from discriminator directs the student to produce output similar to the two vectors above and different from a vector like (0.5, 0.1, 0.4).

\subsubsection{Discriminator update}
We now describe the proposed method in a more rigorous way.  The student and discriminator in Figure \ref{fig:gan} are alternatively updated in the GAN-based approach. Let us first look at the update of the discriminator, which is trained to distinguish teacher and student logits. We use multi-layer perceptron (MLP) as discriminator. Its building block --- residual block is shown in Figure \ref{fig:res} (right). The number of nodes in each layer is the same as the dimension of logits, i.e., the number of categories $C$.
We denote the discriminator that predicts binary value ``Real/Fake'' as $D(\cdot)$. To train $D$, we fix the student network $F(\cdot)$ and seek to maximize the log-likelihood, which is  known as binary cross-entropy loss,
{\small%
\begin{equation}
\hspace{-0.1cm}
\mcL_{A} (D,F) = \frac{1}{N} \sum_{i=1}^N  \Big( \log P(\text{Real} | D(t_i)) + \log P(\text{Fake} | D(F(x_i))) \Big). \label{eq:adv} \nonumber 
\end{equation}
}%
The plain adversarial loss $\mcL_{A}$ for knowledge distillation, which follows the original GAN \cite{goodfellow2014generative}, faces two major challenges. 
First, the adversarial training process is difficult\cite{yadav2017stabilizing}. Even if we replace the log-likelihood with advanced techniques such as Wasserstein GAN \cite{arjovsky2017wasserstein} or Least Squares GAN \cite{mao2016least}, the training is still slow and unstable in our experiments. 
Second, the discriminator captures the high-level statistics of teacher and student outputs, but the low-level alignment is missing. The student outputs $F(x_i)$ for $x_i$ can be aligned to a completely unrelated teacher sample $t_j$ by optimizing $\mcL_{A}$, which means a dog image can generate a logits vector that predicts cat. One extreme example is that the student always mispredicts dog as cat and cat as dog, but the overall output distribution may still be close to the teacher's.

To tackle these problems, we modify the discriminator objective to also predict the class labels, inspired by \cite{chen2016infogan,odena2016conditional}. 
In this case, the output of discriminator $D(\cdot)$ is a $C+2$ dimensional vector with C \emph{Label} predictions and a \emph{Real/Fake} prediction. We now maximize
\begin{equation}
\mcL_{\text{Discriminator}} (D,F) = \frac{1}{2} (\mcL_{A} (D,F) + \mcL_{DS}(D,F)), \label{eq:advfinal}
\end{equation} 
where $\mcL_{A}$ is the previously defined adversarial loss over \emph{Real/Fake}, $\mcL_{DS}$ is the supervised log-likelihood of discriminator over \emph{Labels}, written as
{\small%
\begin{equation}
\mcL_{DS}(D,F) = \frac{1}{N} \sum_{i=1}^N  \Big( \log P(l_i | D(t_i)) + \log P(l_i| D(F(x_i))) \Big). \nonumber
\end{equation} 
}%
We assume \emph{Label} and \emph{Real/Fake} are conditionally independent in Equation \eqref{eq:advfinal}. 
To avoid using this assumption, we can maximize the log-likelihood of discriminator to predict the tuple $\{$ \emph{Label, Real/Fake} $\}$, which requires $D(\cdot)$ to predict a $2C$ dimensional vector. In our experiments, optimizing the GAN-based loss with or without the independent assumption achieves almost identical results. Hence we will always use the independent assumption for a more compact discriminator. \zheng{I think we need a sentence like this to distinguish us from AC-GAN, we are not just one application of AC-GAN, and not necessarily AC-GAN, just coincidence and simplicity}
Note that equation \eqref{eq:advfinal} has the same form as the auxiliary classifier GANs \cite{odena2016conditional}.

The adversarial training becomes much more stable when the proposed discriminator also predicts category \emph{Labels} besides \emph{Real/Fake}.
Moreover, the discriminator can provide category-level alignment between outputs of student and teacher. 
The student outputs of a dog image are more likely to learn from the teacher outputs that predict dogs. 

The GAN-based loss still lacks instance-level knowledge. To exploit the knowledge to further boost the performance,  we start with investigating conditional discriminators, in which the input of discriminators are logits concatenated with a conditional vector. 
We tried the following conditional vectors: image with convolutional embedding; label one-hot vector with embedding; and the extracted teacher logits.
 The embedding includes several weight layers and outputs a vector that is the same size as the logits. 
However, it turns out the conditional vectors are easily ignored during the training of the discriminator. 
The conditional discriminator does not help in practice and we introduce a more direct instance-level alignment for training student network below.

\subsubsection{Student update}
We update the student network after updating the discriminator in each iteration. 
When updating the student network $F(\cdot)$, we aim to fool the discriminator by fixing discriminator $D(\cdot)$ and minimizing the adversarial loss $\mcL_A$. 
In the meantime, the student network is also trained to satisfy the auxiliary classifier of discriminator $\mcL_{DS}$. 
Besides the category-level alignment provided by $\mcL_{DS}$, we introduce instance-level alignment between teacher and student outputs as
\begin{equation}
\mcL_{L_1}(F) = \frac{1}{N} \sum_{i=1}^N \| F(x_i)-t_i \|_1. 
\end{equation} 
The $L_1$ norm has been found helpful in the GAN-based approach for image to image translation \cite{isola2016image}. 

Finally, we combine the learned loss with the supervised loss $\mcL_{S}$ in \eqref{eq:ce}, and minimize the following objective for the student network $F(\cdot)$,
\begin{equation} \label{eq:lossfinal}
\begin{split}
 & \mcL_{\text{Student}}(D,F) = \mcL_{S} (F) + \mcL_{L_1}(F) + \mcL_{GAN}(D,F), 
\\ & \, \text{\small where }  \mcL_{GAN}(D,F) =  \frac{1}{2}(\mcL_{A}(D,F) - \mcL_{DS}(D,F)). 
\end{split} 
\end{equation}
The sign of $\mcL_{DS}$ is flipped in \eqref{eq:advfinal} and \eqref{eq:lossfinal} because both the discriminator and student are trained to preserve the category-level knowledge. 

The final loss $\mcL_{\text{Student}}(D,F) $ in \eqref{eq:lossfinal} is a combination of the learned loss for knowledge distillation and the supervised loss for neural network, and may look complicated at the first glance. 
However, each component of the loss is relatively simple.
 Moreover, since both student $F$ and discriminator $D$ are learned, there is no explicit parameters to be tuned in the loss function. 
 Our experiments in the next section suggest the performance of the proposed method is reasonably insensitive to the discriminator architecture and the learned loss can outperform  the hand-engineered loss for knowledge distillation. 


\section{Experiments}
We present the experimental results in this section. 
The implementation details and experimental settings are provided in section \ref{sec:exp1}.
 We show the benefits of our proposed method compared to knowledge distillation in section \ref{sec:exp2}.
 We then analyze the different loss components of the proposed methods in section \ref{sec:exp3}. 
 The effect of depth and width of the student network is presented in section \ref{sec:exp4}, followed by the discussion of trade-off between classification accuracy and inference time in section \ref{sec:exp5}. 
 Finally in section \ref{sec:exp6}, we show the qualitative visualization on the output distribution for student, teacher, and knowledge distillation. 

\subsection{Experimental setting}
\label{sec:exp1}
We consider three image classification datasets: ImageNet32 \cite{chrabaszcz2017downsampled}, CIFAR-10 and CIFAR-100 \cite{krizhevsky2009learning}. 
ImageNet32 is a downsampled version of the ImageNet2012 challenge dataset \cite{russakovsky2015imagenet}, which contains 1.28M training images and 50K validation images for 1K classes; all images are downsampled to 32$\times$32. 
The CIFAR datasets contain 50K training images and 10K validation images of 10 and 100 classes, respectively. The images are also 32$\times$32. In all the experiments, we perform light data augmentation with horizontal flipping, padding and cropping on input images as in \cite{he2016deep}. 

We use wide residual networks (WRNs) \cite{zagoruyko2016wide} as both student and teacher networks.
 The residual blocks are shown in Figure \ref{fig:res} (right) and the network architectures are in Table \ref{tab:res}. 
WRN-$d$-$m$ denotes network with depth $d$ and widen factor $m$. 
The teacher network is a fixed WRN-40-10, while the student network has varying depth and width in different experiments. Dropout ratio of 0.3 is used for all WRNs. 
We use stochastic gradient descent (SGD) as optimizer, and set the initial learning rate as 0.1, momentum as 0.9, and weight decay as 1e-4. 
For CIFARs, we use minibatch size 128 and train for 200 epochs with learning rate divided by 10 at epoch 80 and 160. For Imagenet32, we use minibatch size 256 and train for 70 epochs with learning rate divided by 10 at epoch 25 and 50. 

We use multi-layer preceptron (MLP) as the discriminator in the GAN-based approach. 3-layer MLP is used for most of the experiments except for section \ref{sec:exp3}, in which we study the effect of discriminator depth. To speed up the experiments, the logits of teacher network are generated offline and stored in memory. For training the discriminator, we use SGD with the same scheduler as in training the student network, but a smaller initial learning rate 1e-3. The logits pass through a batch normalization layer before the MLP. Dropout ratio is also set to 0.3.

The implementation is in PyTorch. The results below are the median of five random runs.  
 
\subsection{Benefits of learning loss}
\label{sec:exp2}

We first show the proposed method is effective for transferring knowledge from teacher to student. Table \ref{tab:base} shows the error rate of classification on the three benchmark datasets. The teacher is  the deep and wide WRN-40-10. The student is much shallower and thinner, WRN-10-4 for CIFARs, and WRN-22-4 for ImageNet32. We choose a larger student network for ImageNet32 because the dataset contains more samples and categories. Section \ref{sec:exp4} and \ref{sec:exp5} have more discussion on wisely choosing the student architecture. 


\begin{table}[tbhp]
\centering
\setlength\tabcolsep{5pt}
\begin{tabular}{|c|c|c|c|}
\hline
 & CIFAR-10 & CIFAR-100 & ImageNet32 \\
\hline
Student & 7.46 & 28.52 & 48.2 \\
\hline
Teacher & 4.19 & 20.62 & 38.41\\
\hline
KD (T=1) & 7.27 & 28.62 & 49.37\\
KD (T=2) & 7.3 & 28.33&49.48\\
KD (T=5) & 7.02 & 27.06 &49.63\\
KD (T=10) & 6.94 & 27.07 &51.12\\
\hline
 Ours & \textbf{6.09} & \textbf{25.75} & \textbf{47.39}\\
\hline
\end{tabular}%
\caption{
Error rate achieved on benchmark datasets. 
}
\label{tab:base}%
\end{table}%

The first two rows of Table \ref{tab:base} show the performance of pure supervised learning for student and teacher networks, without any knowledge transfer. We then compare our GAN-based approach with knowledge distillation (KD) proposed in \cite{hinton2015distilling} and reviewed in section \ref{sec:dk}. We choose the temperature parameter $T \in \{1,2,5,10\}$ following the original work. The GAN-based approach is detailed in section \ref{sec:gan} and no parameter is tuned.

We have several observations from Table \ref{tab:base}. The deep and wide teacher performs much better than the shallow and thin student by pure supervised learning. The error rate of the small network trained with student-teacher strategy is lower bounded by the teacher performance, as expected. Baseline method KD helps the training of small networks for the two CIFARs, but does not help for ImageNet32. We conjecture the reason to be that the capacity of the student is too small to learn from knowledge distillation for larger dataset such as ImageNet32. The temperature parameter $T$ introduced in KD is useful. For CIFARs, KD performs better when $T$ is large, and $T=5$ and $T=10$ performs similarly.  The proposed method improves the performance of small network for all three datasets, and outperforms KD by a margin. 

\subsection{Analysis of the proposed method}
\label{sec:exp3}

\begin{figure}[t]
\centerline{
\includegraphics[width=0.9\linewidth]{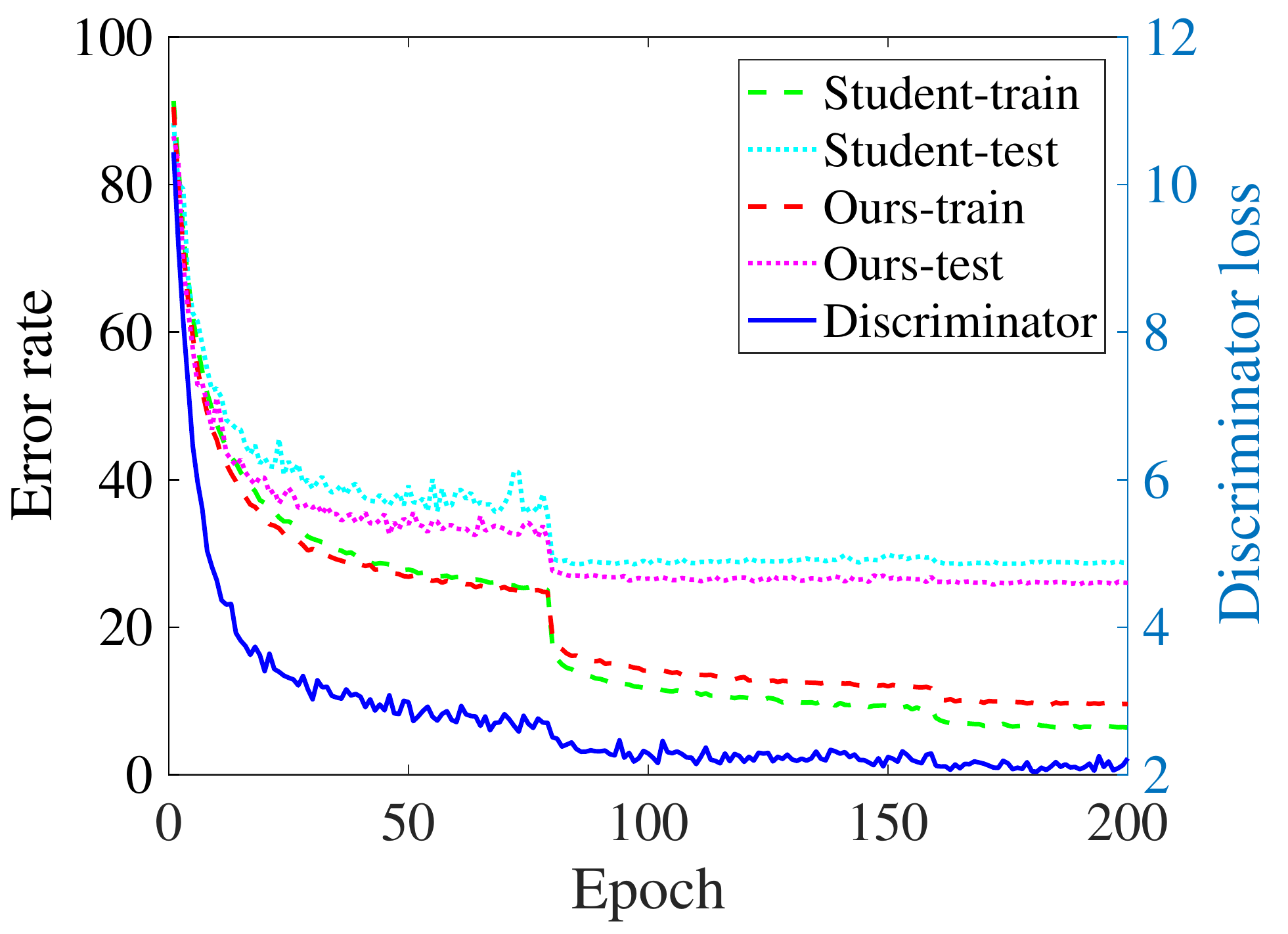}
}
\caption{
The training curve on CIFAR-100. We show the training and testing accuracy of the student network using supervised training and GAN-based training, as well as the discriminator loss.}
\label{fig:train}
\end{figure}

We discuss the proposed method in more detail in this section. Figure \ref{fig:train} presents the training curve of the small student network, WRN-10-4, on CIFAR-100 dataset. The loss of the discriminator (blue solid line) is gradually decreasing, which suggests the adversarial training steadily makes progress. The error rates of GAN-based method for both training and testing data are decreasing. The testing error rate of GAN-based method is consistently better than the pure supervised training of the student model, and looks more stable between epoch 50-100. Surprisingly, the training error rate of the GAN-based method  is slightly worse than pure supervised learning, which suggests knowledge transfer can be more beneficial for generalization. 

\begin{table}[tbhp]
\centering
\begin{tabular}{|c|c|c|}
\hline
Loss composition & CIFAR-10 & CIFAR-100 \\
\hline
$\mcL_S$ & 7.46 & 28.52 \\
\hline
$\mcL_{GAN}$ & 14.82 & 47.04 \\
\hline
$\mcL_S + \mcL_{GAN}$ & 6.56 & 27.27\\
\hline
$\mcL_S + \mcL_{L_1}$  & 6.44 & 26.66 \\
\hline
$\mcL_S + \mcL_{L_1} + \mcL_{GAN}$ & \textbf{6.09} & \textbf{25.75} \\
\hline
\end{tabular}%
\caption{
The effect of different components of the loss in the proposed method; the error rates on CIFARs.
}
\label{tab:comp}%
\end{table}%

Next, we look into the effect of enabling and disabling different components of the GAN-based approach, as shown in Table \ref{tab:comp}. By combining the adversarial loss and the category-level knowledge transfer (Equation \eqref{eq:advfinal}),  the learned loss $\mcL_{GAN}$ performs reasonably well. However, the indirect knowledge provided by $\mcL_{GAN}$ alone is not as good as pure supervised learning $\mcL_S$. Both category-level knowledge transfer by $\mcL_{GAN}$ and instance-level knowledge transfer by $\mcL_{L_1}$ can improve the performance of training student network.  The final approach combines these components and performs the best without parameter tuning.  

\begin{table}[tbhp]
\centering
\begin{tabular}{|c|c|c|c|c|}
\hline
 Depth & 1 & 2 & 3 & 4 \\
\hline
Error rate & 26.13 & 25.88 & \textbf{25.75} & 27.42 \\
\hline
\end{tabular}%
\caption{
The effect of discriminator depth on CIFAR-100.
}
\label{tab:dis-dep}%
\end{table}%

Finally, we present the effect of the depth of MLP as discriminator in Table \ref{tab:dis-dep}. 
The error rate is relatively insensitive to the depth of discriminator. 
The error rate slightly decreases as the depth increases when the discriminator is generally shallow. 
When the discriminator becomes deeper, the error rate increases as the adversarial training becomes unstable. 
Decreasing  the learning rate of discriminator sometimes helps, but it may introduce parameter tuning for the proposed method. 
The 3-layer MLP works reasonably well and is used for all our experiments to keep the GAN-based method simple. 

\begin{figure}[b]
\centering 
		\begin{subfigure}[t]{0.43\textwidth}
                 \includegraphics[width=0.95\textwidth]{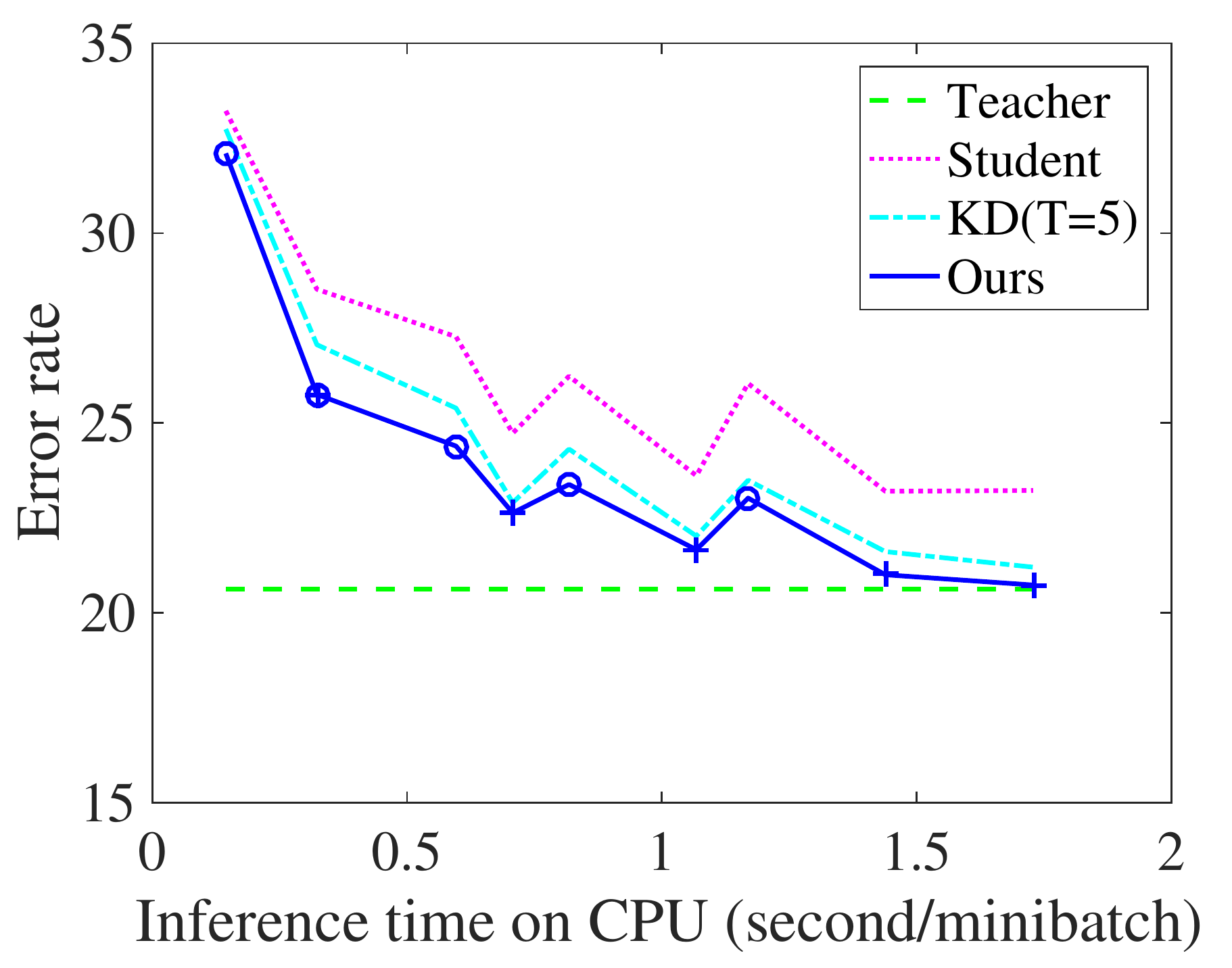}
                 \caption{Trade-off between inference time and error rate.} 
                 \label{fig:time}
         \end{subfigure}
         \\         \vspace{0.3cm} 
         \begin{subfigure}[t]{0.4\textwidth}
                 \includegraphics[width=0.95\textwidth]{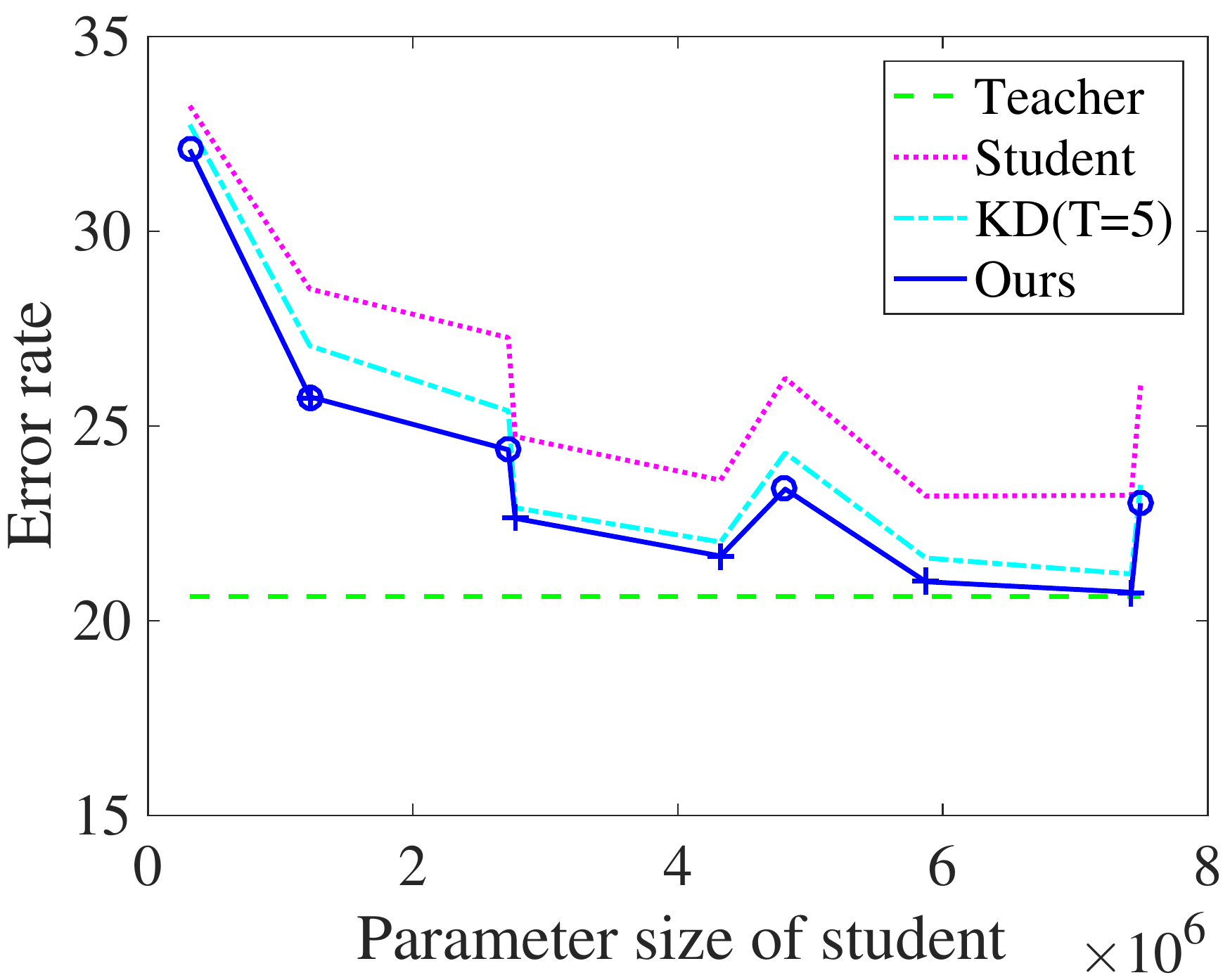}
                 \caption{Trade-off between network size and error rate.} 
					\label{fig:size}
         \end{subfigure}
         \caption{
         Error rate to inference time and parameter size. The figure is generated from Table \ref{tab:wd}.  Networks WRN-10-m are labeled as circles, and WRN-d-4 are labeled as crosses for the GAN-based approach. The largest student is  7x smaller and 5x faster than the teacher WRN-40-10.\zheng{Depth is more effective. The main purpose is to show the trade-off, so I prefer 1 plot for each method. I agree splitting depth and width will make the figure smooth, but it'll be complicated with several lines and hard to address our points. We already discussed depth and width in the previous section.}
} \label{fig:tisi}
\end{figure}

\subsection{Does WRN need to be deep and wide?}
\label{sec:exp4}

\begin{table}[tbhp]
\centering
\setlength\tabcolsep{3.5pt}
\begin{tabular}{|c|c|c|c|c|>{\bfseries}c|}
\hline
 WRN & Size (M) & Time (s) & Student & KD (T=5) & Ours \\
\hline
10-2 &0.32 &0.14 & 33.22 & 32.74 &  32.1\\
10-4 &1.22 &0.32 & 28.52 &  27.16 & 25.75 \\
10-6 &2.72 &0.60 & 27.27 &  25.39 & 24.39   \\
10-8 &4.81 &0.82 & 26.23 & 24.31 & 23.38 \\
10-10 &7.49 &1.17 & 26.04 &  23.49 & 23.02  \\
\hline
16-4  &2.77 & 0.71 & 24.73 & 22.9 & 22.73\\
22-4 &4.32 &1.07 & 23.61 & 22.02 & 21.66\\
28-4 &5.87 &1.44 & 23.2 & 21.61 & 21.00 \\
34-4 &7.42 &1.73 & 23.22 & 21.2 & 20.73\\
\hline
40-10 &55.9 &8.73 & 20.62 & - & -\\
\hline
\end{tabular}%
\caption{
The effect of depth and width in student network; the parameter size, inference time and error rate on CIFAR-100.
}
\label{tab:wd}%
\end{table}%

\cite{urban2016deep} asked similar question for convolutional neural networks and claimed the network should at least has a few layers of convolutions.  
We study the modern architecture WRN of residual blocks. 
Our empirical study suggests that even for the modern architecture WRN, the network has to be deep and wide to some extent.



\begin{figure*}[t]
\centerline{
\includegraphics[width=0.9\linewidth]{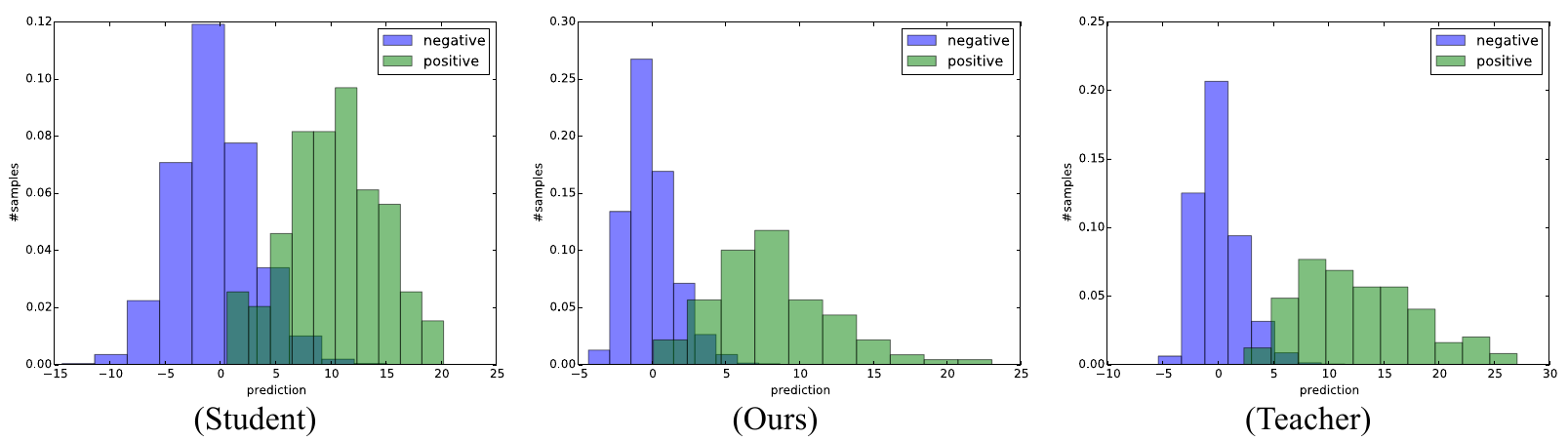}
}
\caption{
Qualitative visualization; the distribution of prediction for category 85 in CIFAR-100.\zheng{this is the histogram of prediction value. It may help to use same x/y-axix, but the data is stored in Honda server and I cannot access. Only want to show qualitative results.}}
\label{fig:vis}
\end{figure*}

Table \ref{tab:wd} presents the results of pure supervised learning, knowledge distillation \cite{hinton2015distilling} and the GAN-based approach for different student networks on CIFAR-100. 
We first fix the depth of WRN as 10, and change the widen factor from 2 to 10. 
10 is the minimum depth for our WRN architecture as the depth has to be $6n+4$. 
We then fix the width as 4, and increase depth from 10 to 34. 
The parameter size is in millions, and the inference time is in seconds per minibatch of 100 samples on CPU. 

When the student is very small, such as WRN-10-2, it is difficult to transfer knowledge from teacher to student because the student is limited by its network capacity.  
When the student is large, such as WRN-34-4, both knowledge distillation and GAN-based approach can improve the performance close to the level of the teacher. 
The advantage of the proposed method is observed at all depths and widths but is most pronounced for relatively small students such as WRN-10-4.  
Increasing depth is more effective than increasing width for WRN. For example, WRN-34-4 has less parameter than WRN-10-10, but achieves lower error rate. 

\subsection{Training student for acceleration}
\label{sec:exp5}

%

The shallow and thin network is much easier to deploy in practice. We present the trade-off between error rate, inference time and parameter size in Figure \ref{fig:tisi}. 
The figure is generated from Table \ref{tab:wd} by changing the architecture of the student network. 
Larger student network is more accurate but also slower. 
For network with similar size, such as WRN-10-10 and WRN-34-4, deeper network achieves lower error rate, while wider network runs slightly faster. 
The student-teacher strategy can help improve the classification performance of the student network. 
When the student network is relatively large, such as WRN-34-4, the student network trained by the GAN-based approach can achieve error rate comparable to the teacher WRN-40-10, while being 7x smaller and 5x faster. 
Compared to the baseline student by pure supervised training, the GAN-based approach decreases the absolute error rate by 2.5\%.

\subsection{Visualization of distribution}
\label{sec:exp6}

In the last section of experimental results, we present  qualitative visualization for the GAN-based approach. Figure \ref{fig:vis} presents the scaled histogram for the prediction of category 85 in CIFAR-100. The histogram is calculated on the 10K testing samples, in which 100 samples are from category 85 and labeled as positive (green in figure), and the other 9.9K are labeled as negative (blue in the figure). The histogram is normalized to sum up to one for positive and negative, respectively. The three plots represent the distribution predicted by student network trained by pure supervised learning, the student network trained by GAN-based approach, and the teacher network. 
The histogram in the middle is similar to the histogram on the right, which suggests the GAN-based approach effectively transfers knowledge from teacher to student. 

\section{Conclusion and discussion}
We study the student-teacher strategy for network acceleration in this paper. 
We propose a GAN-based approach to learn the loss for transferring knowledge from teacher to student. 
We show that the GAN-based approach can improve the training of student network, especially when the student network is shallow and thin. 
Moreover, we empirically study the effect of network capacity when adopting modern network as student and provide guidelines for wisely choosing a student to balance error rate and inference time. 
In specific settings, we can train a student that is 7x smaller and 5x faster than teacher without loss of accuracy.  

The GAN-based approach is stable and easy to implement after applying several advanced techniques in the GAN literature. The current implementation uses the stored logtis from teacher network to save GPU memory and computation. Generating teacher logits on the fly with dropout can be more reliable for the adversarial training. At last, the GAN-based approach can be naturally extended to use ensemble of networks as teacher. The logits of multiple teacher networks can be fed into the discriminator for better performance. We will investigate these ideas for future work.


{\small
\balance
\bibliographystyle{ieee}
\bibliography{net,acc,gan}
}

\end{document}